\documentclass{article} % For LaTeX2e
\usepackage{iclr2027_conference,times}

\usepackage{graphicx}
\usepackage{amsmath}
\usepackage{amssymb}
\usepackage{booktabs}
\usepackage{array}
\usepackage{multirow}
\usepackage{adjustbox}
\usepackage{subcaption}
\usepackage{xspace}
\usepackage{xcolor}
\usepackage{pifont}
\usepackage{caption}
\usepackage{wrapfig}

\usepackage{hyperref}
\usepackage{url}
\hypersetup{
    pdftitle={StoryEngine: A State-Grounded Agentic Framework for Video Storytelling},
    pdfauthor={Yingrui Wang, Zeqing Wang, Yeying Jin}
}

\usepackage{float}
\usepackage{listings}
\DeclareCaptionStyle{ruled}{labelfont=normalfont,labelsep=colon,strut=off}
\floatstyle{ruled}
\newfloat{listing}{tb}{lst}
\floatname{listing}{Listing}

\newcommand{\method}{\text{StoryEngine}\xspace}

\title{\raggedright StoryEngine: A State-Grounded Agentic Framework for Video Storytelling}

\author{%
Yingrui Wang$^{1,2\dagger}$, \quad
Zeqing Wang$^{1,2\dagger}$, \quad
Yeying Jin$^{1,2\ddagger *}$ \\
{\normalfont $^{1}$Tencent} 
{\normalfont $^{2}$National University of Singapore} \\
{\normalfont\texttt{yingrui\_wang@u.nus.edu}, \texttt{zeqing.wang@u.nus.edu},} \\
{\normalfont\texttt{jinyeying@u.nus.edu}}
}

\iclrpreprintcopy % Public preprint with named authors and no review line numbers.

\begin{document}

\maketitle

% Author contribution notes follow the reference paper's first-page footnotes.
\begingroup
\renewcommand{\thefootnote}{\fnsymbol{footnote}}
\footnotetext[2]{Equal contribution.}
\footnotetext[3]{Project lead.}
\footnotetext[1]{Corresponding author.}
\endgroup

\begin{abstract}

Despite recent progress in agentic multi-shot video generation, producing coherent and consistent long-form stories remains challenging. Existing agentic pipelines typically rely on textual shot plans or previously generated pixels, yet lack an explicit mechanism for propagating the consequences of story events and maintaining the video world state across shots. As a result, missing visual details may be reconstructed inaccurately, while visual drift may propagate across subsequent shots, undermining both narrative coherence and visual consistency.
To address these challenges, we propose \textbf{\method}, a state-grounded agentic framework for video storytelling. \method establishes a separation between authoritative semantic plans and unreliable visual observations. Specifically, \method maintains a structured representation of entity placement and story-relevant states, and propagates event-induced changes to define the intended start and end states of each shot. To visually realize these states, \method constructs canonical references for recurring entities and environments, and compiles state and visual constraints into executable render plans. 
A bounded, evaluation-guided repair loop further ensures that each shot faithfully depicts its intended states by correcting residual inconsistencies. Together, these mechanisms preserve causal story progression and prevent local visual errors from propagating across shots.
To comprehensively evaluate long-form storytelling, we construct a benchmark across diverse scenarios and visual styles, with metrics assessing storytelling quality, narrative coherence, and visual consistency. Experimental results demonstrate that \method consistently outperforms state-of-the-art methods across all evaluation dimensions, validating its effectiveness for coherent and consistent video storytelling. Project page: \url{https://wwwtaylor.github.io/StoryEngine/}. 

\end{abstract}

\section{Introduction}
Modern video generators~\citep{Wan,veo,seedance} can synthesize visually compelling short clips from text or images. Long-form storytelling~\citep{storydiffusion,videoinfinity}, however, requires more than a collection of individually plausible shots: it demands a sequence of interconnected shots that collectively advance a coherent and consistent narrative. Achieving such cross-shot coordination manually requires substantial human effort to plan, generate, and supervise each shot.

To reduce this burden, recent agentic systems~\citep{movieagent,vimax,codirector} automate video production by decomposing it into specialized stages, such as scriptwriting, storyboarding, asset creation, and shot-level generation. By coordinating these stages under a shared narrative plan, such systems can organize multi-shot sequences and improve cross-shot visual consistency with less human intervention.

Despite this progress, existing agentic systems still struggle to preserve coherent and consistent story progression. Many pipelines~\citep{vimax,movieagent,a2rd} represent narratives using loosely structured textual shot descriptions while conditioning later shots on previously generated frames or clips. This formulation remains insufficient in three respects. \textit{First}, textual descriptions specify what should happen, but do not explicitly track how story events update entity placements and story-relevant attributes across shots. \textit{Second}, they provide limited support for translating these states into shot-specific visual constraints: recurring entities and environments must remain identifiable. \textit{Third}, because generated pixels are repeatedly reused as context, a misplaced object or a corrupted detail may be interpreted as a new narrative fact rather than a local execution error. Semantic intent thus becomes entangled with fallible visual observations, allowing local generation failures to accumulate into global narrative incoherence and cross-shot visual inconsistencies.

\begin{figure}[t]
    \centering
    \includegraphics[width=0.9\textwidth]{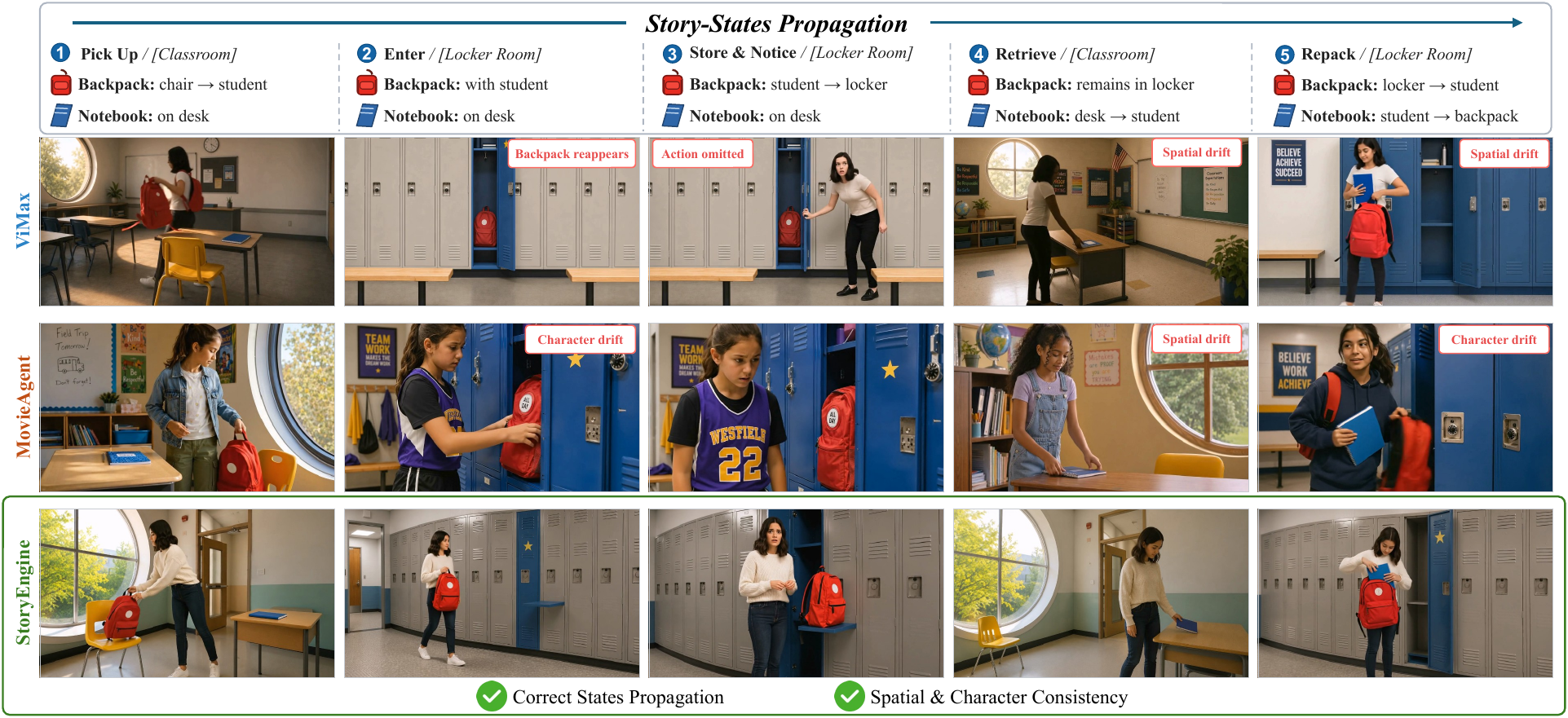} 
    \caption{A multi-shot video comparison among ViMax, MovieAgent, and \method. The story follows a student retracing her steps to recover a forgotten notebook and reunite it with her backpack. \method correctly propagates story states across shots (e.g., those of the backpack and notebook) while maintaining spatial and character consistency.}
    \label{fig:teaser}
\end{figure}

Consider the story illustrated in Figure~\ref{fig:teaser}, in which a student stores her backpack in a locker, returns to the classroom for a forgotten notebook, and then revisits the locker room. Although simple, the sequence requires the consequences of earlier events to persist: the backpack must remain in the locker while the student is in the classroom, and the notebook must remain on the desk until it is collected. Meanwhile, the recurring classroom, locker room, and objects must preserve their visual identity, and the selected views must clearly reveal the placement and retrieval actions. If a generated shot omits an action or violates one of these states, the system should be able to repair the local output. Good storytelling therefore requires explicit state propagation, visually grounded render planning, and a repair process constrained by the intended semantic plan.

To meet these requirements, we propose \textbf{\method}, a state-grounded agentic framework for coherent video storytelling. At its core, \method{} separates an \emph{authoritative semantic plan} from \emph{fallible visual observations}. Operationally, \method organizes generation into three stages: story-state propagation makes the intended evolution of the story world explicit; grounded render planning translates this evolution into generator-ready visual constraints; and continuity-aware generation realizes each shot while preventing visual errors from feeding back into the story. 

Specifically, \emph{story-state propagation} represents recurring entities through their placements and story-relevant attributes, converts narrative events into explicit state transitions, and propagates these changes to determine the intended start and end states of each shot. As a result, the opening state of a shot follows the planned story progression rather than being inferred from potentially erroneous generated pixels. \emph{Grounded render planning} then constructs canonical references for recurring entities and scenes, selects action-relevant camera views, and compiles state, camera, and visual constraints into executable render plans that specify how each transition should be depicted. Finally, \emph{continuity-aware generation} evaluates preceding visual evidence against the intended opening state of the next shot, reusing, selectively referencing, or discarding it as appropriate. A bounded evaluation-guided repair loop further corrects local state inconsistencies while keeping the authoritative semantic plan fixed. In this way, compatible visual evidence improves cross-shot continuity, whereas omitted actions, misplaced objects, and corrupted details remain local generation errors rather than being propagated as narrative facts.

Together, these stages determine what must be true, how it should be shown, and which visual evidence is safe to carry forward. Experiments on a 60-story benchmark with two video backbones demonstrate that \method consistently outperforms state-of-the-art agentic systems on all metrics of narrative realization, cross-shot coherence, and visual consistency.

In summary, our contributions are as follows:
\begin{itemize}
    \item We propose \method, a state-grounded agentic framework that separates authoritative semantic plans from fallible visual observations and explicitly propagates story states across shots.
    
    \item We develop grounded render planning and continuity-aware generation to translate propagated states into executable visual constraints and to repair local errors.
    
    \item We introduce a 60-story benchmark with three diagnostic suites and frozen annotations to evaluate narrative realization, cross-shot coherence, and visual consistency. Experimental results show that \method outperforms representative baselines on all metrics.
\end{itemize}

\section{Related Work}
\label{sec:related_work}

\paragraph{Video generation.}
Modern short-form video models~\citep{hunyuanvideo,sora}, such as Wan~\citep{Wan}, Veo~\citep{veo}, and Seedance~\citep{seedance}, can synthesize visually compelling clips from text or images. Long-form storytelling, however, requires coherent and visually consistent shots that collectively convey a narrative. Recent methods extend generation across multiple shots through shared visual context, longer temporal modeling, or reference frames~\citep{shotadapter,longcontext,stage,dualparal}. Nevertheless, their per-shot quality often lags behind state-of-the-art short-form models, while cross-shot narrative coherence and visual consistency remain challenging.

\paragraph{Agentic video generation.}
To leverage the strong capabilities of short-form video models for long-form storytelling, traditional production pipelines rely heavily on human creators to plan narrative progression, specify individual shots, and verify narrative and visual consistency.
Agentic systems reduce the manual effort of long-form production by decomposing it into specialized stages. StoryAgent coordinates agents for planning, storyboarding, generation, and evaluation~\citep{storyagent}, while MovieAgent employs hierarchical director and shot-planning agents~\citep{movieagent}. Recent systems further combine global planning with iterative refinement or propagate visual anchors across shots~\citep{codirector,vimax}. 
Although these methods automate multi-shot production, they typically coordinate shots through textual descriptions, retrieved context, or generated images. Such representations describe what should occur, but do not explicitly propagate event consequences or distinguish intended story facts from errors in generated pixels. Consequently, omitted actions, misplaced objects, and corrupted details may be carried into later shots. In contrast, \method separates authoritative semantic plans from fallible visual observations, explicitly propagates story states, and realizes them through grounded render planning and bounded local repair.

\section{Method}
\label{sec:method}

\begin{figure}[t]
    \centering
    \includegraphics[width=0.9\textwidth]{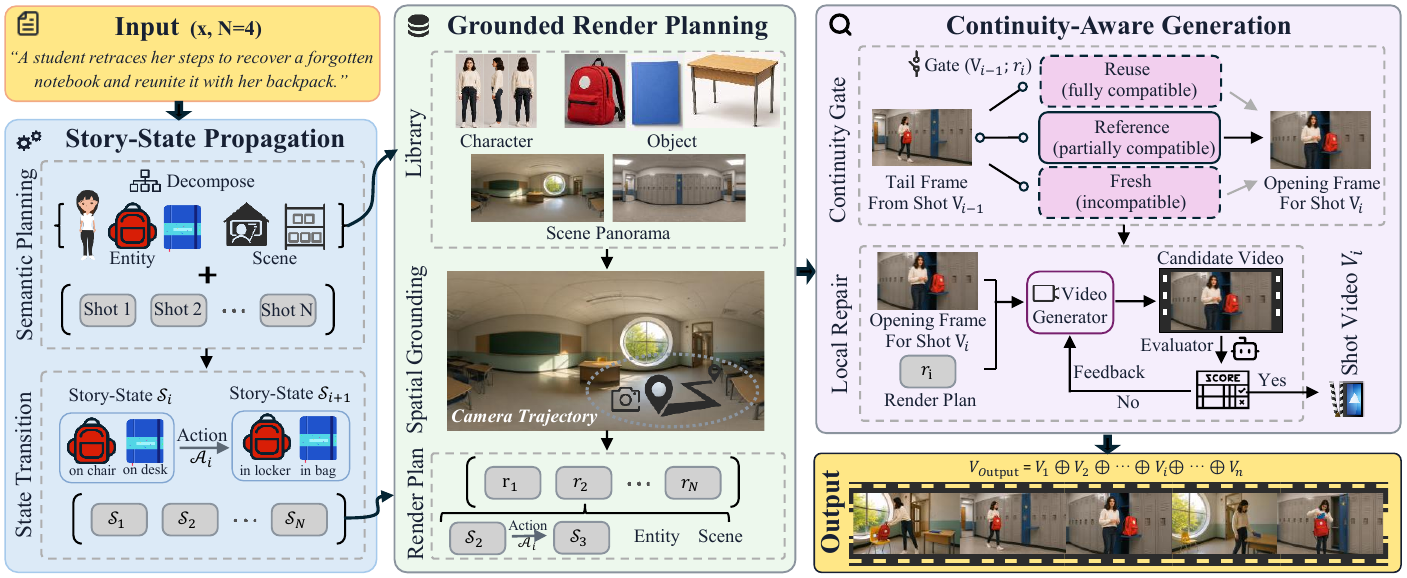}
    \caption{Overview of \method{}. Given a story request \(x\) and a target of
    \(N\) shots, the framework proceeds from left to right through three stages.
    Story-state propagation decomposes the request into entities, scenes, 
    and shots, and tracks action-conditioned transitions between semantic states.
    Grounded render planning constructs canonical character, object, and scene
    references together with camera constraints, yielding shot-level render
    plans. Continuity-aware generation admits prior visual evidence only when
    compatible and applies evaluation-guided local repair before concatenating
    the accepted shots into the final video \(\mathcal{V}\).}
    \label{fig:overview}
\end{figure}

\subsection{Overview}
\label{sec:method_overview}
Given a story request \(x\) and a target number of shots \(N\), our goal is to generate a long-form video \(\mathcal{V}=v_1\oplus\cdots\oplus v_N\), where \(v_i\) is shot \(i\) and \(\oplus\) denotes temporal concatenation. The central challenge is preserving story-event consequences across shots without treating imperfect pixels as narrative truth. To address this, \method{} separates an \emph{authoritative semantic plan} from \emph{fallible visual observations}. The semantic plan defines story truths and remains fixed during generation. Generated images, videos, and tail frames are observations that help realize, but cannot revise, the plan.

The complete framework is
\begin{equation}
    \begin{aligned}
        \mathcal{P}_{\mathrm{s}}
            &= F_{\mathrm{state}}(x,N),
        &\quad (\mathcal{L},\mathcal{U})
            &= F_{\mathrm{ground}}(\mathcal{P}_{\mathrm{s}}),\\
        \mathcal{P}_{\mathrm{r}}
            &= C_{\mathrm{render}}(\mathcal{P}_{\mathrm{s}},\mathcal{L},\mathcal{U}),
        &\quad c_i
            &= \mathrm{Gate}(v_{i-1};r_i),\quad r_i\in\mathcal{P}_{\mathrm{r}}.
    \end{aligned}
    \label{eq:storyengine_overview}
\end{equation}

First, story-state propagation \(F_{\mathrm{state}}\) converts \(x\) and \(N\) into the authoritative semantic plan \(\mathcal{P}_{\mathrm{s}}\), which specifies recurring entities and scenes, the initial world state, per-shot events, spatial intents, and story requirements.

Next, grounded render planning uses \(F_{\mathrm{ground}}\) to construct the canonical reference library \(\mathcal{L}\) and grounded contexts \(\mathcal{U}=(\mathcal{U}_1,\ldots,\mathcal{U}_N)\). Each \(\mathcal{U}_i\) provides the scene view, camera, and reference bindings for shot \(i\). The compiler \(C_{\mathrm{render}}\) combines them with \(\mathcal{P}_{\mathrm{s}}\) into the executable render plan \(\mathcal{P}_{\mathrm{r}}=(r_1,\ldots,r_N)\), where each contract \(r_i\) binds the propagated states and criteria of shot \(i\) to \(\mathcal{U}_i\).

Finally, continuity-aware generation treats the selected tail frame of \(v_{i-1}\) as an observation. The compatibility resolver \(\mathrm{Gate}\) checks it against \(r_i\) and returns the continuity input \(c_i\) in \emph{reuse}, \emph{reference}, or \emph{fresh} mode. Conditioned on \(r_i\) and \(c_i\), \(F_{\mathrm{gen}}\) obtains the opening frame, generates \(v_i\), and invokes bounded evaluation-guided repair.

\subsection{Story-State Propagation}
\label{sec:story_state_propagation}

\paragraph{Structured semantic plan.}
Within \(F_{\mathrm{state}}\), a planning model decomposes the request into recurring entities, environments, and an ordered sequence of shots. The resulting \(\mathcal{P}_{\mathrm{s}}\) contains the recurring entity set \(\mathcal{E}\), an environment catalog, and the initial world state \(\mathcal{S}_1\). For every shot \(i\in\{1,\ldots,N\}\), it also contains an ordered event list \(\mathcal{A}_i\) and a spatial intent \(\eta_i\), together with shot-level requirements and cross-shot ordering constraints. Each event couples a narrative action with an optional effect on the world, while \(\eta_i\) describes how the events should be presented.

A world state $\mathcal{S}$ maps every recurring entity $e\in\mathcal{E}$ to a placement and its story-relevant attributes, i.e., $\mathcal{S}(e)=(\text{placement},\text{attributes})$.
Placements include both scene locations and relations such as being held by a character, placed on a surface, or stored inside a container. Attributes capture conditions whose changes matter to the story or its visual realization, such as whether an object is open, broken, or empty. This representation is therefore compact: it models the facts needed for causal and visual continuity, rather than attempting to reconstruct the entire physical world.

\paragraph{Event-induced state transitions.}
For shot \(i\), let \(\mathcal{S}_i\) and \(\mathcal{S}_{i+1}\) denote the intended states immediately before and after the shot. The deterministic state reducer \(R\) applies the ordered event effects in \(\mathcal{A}_i\), yielding the next state $\mathcal{S}_{i+1}=R(\mathcal{S}_i,\mathcal{A}_i)$.

Internally, the reducer processes events in narrative order; events without state-changing effects leave the state unchanged. It validates entity references, placement relations, and attribute domains after each update. The pair \((\mathcal{S}_i,\mathcal{S}_{i+1})\) thus forms the semantic state contract for shot \(i\).

Crucially, \(\mathcal{S}_{i+1}\) follows the planned effects in \(\mathcal{A}_i\), not an estimate recovered from the generated video \(v_i\). Missed or corrupted details therefore remain local execution errors rather than becoming new narrative facts for subsequent shots.

\subsection{Grounded Render Planning}
\label{sec:visual_grounding}

The semantic state contract determines \emph{what} must be true, but not how those facts should appear in pixels. Grounded render planning supplies the missing identity and spatial evidence, and compiles it with the propagated states into generator-ready constraints.

\paragraph{Canonical visual grounding.}
For each recurring entity, \method{} constructs a canonical identity reference; when a story-relevant attribute changes its appearance, separate state-specific references are created. For each recurring environment, the framework builds a canonical panoramic reference that anchors its appearance and semantic layout. Candidate assets are evaluated against identity or layout criteria before selection. The selected assets form the shared library \(\mathcal{L}\) and serve as visual anchors rather than semantic state. Consequently, visual conditioning is selected from the planned state trajectory instead of being regenerated from the wording of each shot.

Together with \(\mathcal{L}\), \(F_{\mathrm{ground}}\) constructs the grounded contexts \(\mathcal{U}=(\mathcal{U}_1,\ldots,\mathcal{U}_N)\). For shot \(i\),
\(\mathcal{U}_i\) packages an action-relevant scene view and camera specification, together with bindings to state-appropriate assets. The spatial intent \(\eta_i\) identifies the action region, story-relevant landmarks, framing, and directional constraints. Grounding these requirements in a shared environment reference encourages independently generated views to preserve scene identity while clearly revealing the planned interaction.

\paragraph{Executable render plan.}
Once the required visual evidence is available, \(C_{\mathrm{render}}\) combines \(\mathcal{P}_{\mathrm{s}}\), \(\mathcal{L}\), and \(\mathcal{U}\) into shot-level contracts. Let \(\mathcal{K}_i\) denote the resulting set of phase-tagged criteria for shot \(i\). The rendering contract and complete render plan are
\begin{equation}
    \begin{aligned}
        r_i &= (\mathcal{S}_i,\mathcal{A}_i,
                 \mathcal{S}_{i+1},\mathcal{U}_i,\mathcal{K}_i),\\
        \mathcal{P}_{\mathrm{r}} &= (r_1,\ldots,r_N).
    \end{aligned}
    \label{eq:executable_render_shot}
\end{equation}
Criteria tagged as \emph{start}, \emph{motion}, and \emph{end} are compiled, respectively, from visible facts in \(\mathcal{S}_i\), the events in \(\mathcal{A}_i\), and visible facts in \(\mathcal{S}_{i+1}\); applicable requirements are added as phase-specific or always-on criteria. Thus, \(r_i\) jointly specifies the intended opening composition, the transition that must occur, and the evidence required at the end of the shot. It translates semantic and visual constraints into an executable interface without asking the generator to infer persistent story facts from free-form text or preceding pixels.

\subsection{Continuity-Aware Generation}
\label{sec:continuity_aware_generation}

Continuity-aware generation connects consecutive shots through a one-way
evidence gate. Preceding pixels may condition the next shot when they are
compatible with its intended opening state, but cannot modify the propagated
state trajectory or its compiled criteria.

\paragraph{Continuity gate.}
Shots are generated in temporal order. Before rendering shot \(v_i\), the system evaluates the preceding observation \(v_{i-1}\), represented by the selected tail frame of shot \(i-1\), against the start-tagged criteria and visual context in \(r_i\); the first shot has no preceding observation. Formally,
\begin{equation}
    c_i = \mathrm{Gate}(v_{i-1};r_i),
    \label{eq:first_frame_modes}
\end{equation}
where \(c_i\) packages one of three modes---\emph{reuse}, \emph{reference}, or \emph{fresh}---together with any admissible visual evidence or guidance. In reuse mode, a compatible tail serves directly as the opening frame. In reference mode, only compatible content is retained when composing a new opening frame, with inconsistent content explicitly excluded. In fresh mode, an unavailable or incompatible tail is discarded. The resulting opening frame is checked against the start-tagged criteria before video generation; a rejected reuse proposal falls back to reference or fresh mode. This adaptive policy preserves useful cut-to-cut continuity without allowing a
misplaced object or an omitted state change to propagate forward.

The initial candidate from \(F_{\mathrm{gen}}\) is
\begin{equation}
    v_i^{(0)}
    = F_{\mathrm{gen}}(r_i,c_i).
    \label{eq:shot_video_generation}
\end{equation}

\paragraph{Plan-preserving local repair.}
Because a valid render plan does not guarantee a correct sample, \method{} uses a bounded evaluation-guided repair loop. It employs a VLM~\citep{gemini35flash} as the evaluator to judge each candidate video. At repair iteration \(t\), let \(\mathcal{B}_i^{(t)}\subseteq\mathcal{K}_i\) be the criteria evaluated as \(\mathrm{FAIL}\) or \(\mathrm{UNKNOWN}\), and let \(T\) be the fixed repair budget. The next candidate is generated as
\begin{equation}
    v_i^{(t+1)}
    = F_{\mathrm{gen}}\!\left(r_i,c_i;\mathcal{B}_i^{(t)}\right),
    \quad 0\leq t<T.
    \label{eq:local_repair}
\end{equation}
The correction uses the original canonical statements associated with \(\mathcal{B}_i^{(t)}\), while \(r_i\) and \(\mathcal{P}_{\mathrm{s}}\) remain unchanged. The first candidate satisfying all required criteria is returned as \(v_i\). If the budget is exhausted, the system instead selects as \(v_i\) the technically valid candidate that best satisfies state, action, identity, and continuity requirements, and records any remaining violation as a local degradation.

Consequently, the only authoritative semantic link across shots is the planned transition from \(\mathcal{S}_i\) to \(\mathcal{S}_{i+1}\). The selected video \(v_i\) is supplied for continuity, but never feeds back into the story state. This separation preserves causal progression while preventing local visual failures from accumulating across the long-form video.
\section{Experiments}\label{sec:experiments}
\subsection{Benchmark}\label{sec:benchmark}
\begin{wrapfigure}{r}{0.44\textwidth}
    \centering
    \vspace{-1.2em}
    \includegraphics[width=0.44\textwidth]{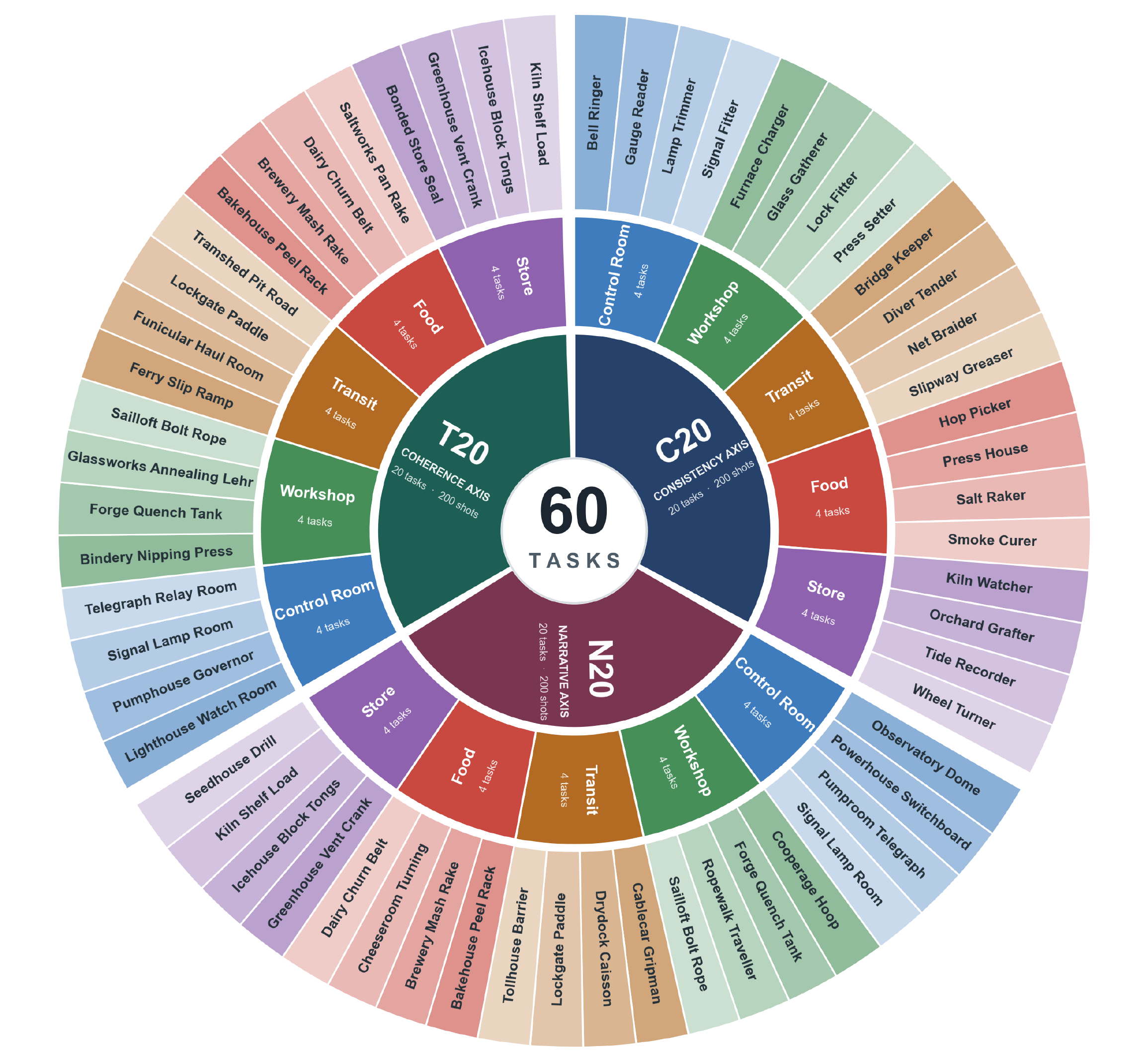}
    \vspace{-1.8em}
    \caption{Benchmark composition (inner to outer ring): evaluation suites, setting categories, and stories.}
    \label{fig:benchmark}
    \vspace{-1.0em}
\end{wrapfigure}
We construct a benchmark of 60 multi-shot stories. The evaluation targets are fixed before generation, so that failures can be localized to individual shots and cuts. As shown in Figure~\ref{fig:benchmark}, it comprises three diagnostic suites of 20 stories, each targeting one evaluation dimension. Each suite contains four stories from each of five setting categories (Store, Food, Transit, Workshop, and Control Room). Each story requests 10 shots over two recurring locations. Its brief, the only input given to a method, specifies the story idea, shot-to-location plan, recurring entities, visual style, and forbidden content. Each brief is paired with an annotation sheet that is hidden from all methods, authored together with the brief without consulting any system output, and frozen under a content hash.

\textbf{N20} targets narrative realization. Each brief lists 10 events that must happen on screen as visible changes, and the annotation sheet adds six distractor events drawn from other N20 stories to expose false-positive judgments. \textbf{T20} targets cross-shot coherence. For each of the nine cuts in a story, the brief names a large, distinctive anchor entity that must be visible on both sides of the cut; 71 of the resulting 180 anchored cuts coincide with a change of location. Each T20 story also specifies three irreversible state changes, which the system may schedule freely but must never undo. \textbf{C20} targets visual consistency. Each story features a single character, and its shot plan switches between the two locations five or six times; in 13 stories, the visual style additionally prescribes a lighting contrast between them. Appendix~\ref{app:benchmark} provides per-suite statistics, the complete story index, and example briefs.

\subsection{Evaluation Metrics}
\label{sec:evaluation_protocol}

We evaluate narrative realization, cross-shot coherence, and visual consistency with eight complementary metrics (higher is better for all). PEC and ECS are computed on N20; APR, SPS, and LAR on T20; GPC and LCS on C20; and MIR on all 60 stories. Gemini-3.5-Flash~\citep{gemini35flash}, queried at temperature 0, serves as the vision-language model (VLM) judge: it scores PEC, ECS, APR, SPS, and LAR and provides the location gate of GPC, whereas MIR and LCS involve no VLM. Appendix~\ref{app:evaluation} gives the complete definitions, frame sampling, scoring rules, and thresholds.

\noindent \textbf{Plan Event Coverage (PEC).}
PEC measures narrative coverage before rendering, whether the per-shot plan explicitly states each required event. A match counts as clear only if it is supported by a quotation verified against the shot text. Events stated in more than half of the shots receive only partial credit, the score is corrected for the credit assigned to distractor events, so that indiscriminate matching does not inflate coverage. Because PEC evaluates the story plan, it is not applicable to methods without a story-planning module; their entries are marked with a dash (--) in the tables.

\noindent \textbf{Event Completion Score (ECS).}
ECS measures whether the required events visibly occur in the generated video. An event counts as clearly completed only if the judge cites two frames of the same shot that show the change; a static depiction of the event's participants or outcome counts only as partial evidence. The false-positive rate on distractor events is subtracted from the score.

\noindent \textbf{Anchor Persistence Rate (APR).}
APR measures whether the designated anchor entity is visible on both sides of each planned cut, judged from frames in the last 0.4~s of the outgoing shot and the first 0.4~s of the incoming shot. Changes in camera angle, framing, or lighting are not counted as violations, whereas cuts with unreadable evidence count against the score.

\noindent \textbf{State Progression Score (SPS).}
SPS measures whether each annotated irreversible state change progresses from its initial to its final state without later reversal. It orders the state evidence from all shots planned at the state's location, rewarding a single forward transition and penalizing sequences that lack one of the two states or revert to an earlier one.

\noindent \textbf{Location Adherence Rate (LAR).}
LAR measures whether each shot depicts its planned location, computed as the fraction of shots with readable location evidence whose judged location matches the plan.

\noindent \textbf{Geometric Place Consistency (GPC).}
GPC measures whether shots planned at the same location depict geometrically compatible backgrounds. For each such shot pair, it masks out people, verifies background feature correspondences with robust two-view geometry, and requires the judged locations of the two shots to agree. GPC thus complements LAR by testing scene geometry across views in addition to location identity.

\noindent \textbf{Minimum Identity Retention (MIR).}
MIR measures facial identity consistency across shots. It groups face embeddings into identity clusters and reports the mean cross-shot similarity of the least consistent cluster, so that a highly consistent identity cannot compensate for a less consistent one.

\noindent \textbf{Lighting Coherence Score (LCS).}
LCS measures whether lighting is consistent within each annotated lighting condition and distinct across a declared contrast. It compares photometric similarity within and between conditions, so that both a uniform look throughout the story and lighting changes that ignore the plan receive low scores.

\subsection{Experimental Setup}\label{sec:experimental_setup}
\paragraph{Baselines.}
We compare \method{} with two agentic systems, ViMax~\citep{vimax} and MovieAgent~\citep{movieagent}, and with Direct I2V, a planner-free baseline that parses each brief into per-shot prompts (location, opening anchor, entities, and style) and generates every keyframe and shot independently. All methods receive the same briefs and are run with two video backbones, Veo~3.1~\citep{veo} and Wan2.2-TI2V-5B~\citep{Wan}; within each comparison, all methods use the same backbone.

\paragraph{Implementation details.}
All LLM-based methods plan with GPT-5.5~\citep{gpt55}, and all methods generate images with GPT-Image-2~\citep{gptimage2}. \method{} uses Gemini-3.5-Flash~\citep{gemini35flash} as its in-loop evaluator with a repair budget of $T=3$. Appendices~\ref{app:implementation} and~\ref{app:baselines} give the remaining settings and the configurations of all baselines and variants.

\subsection{Main Results} \label{sec:main_results}
\begin{table}[!t]
\caption{Main results with two video backbones. Avg is the unweighted mean of the seven video-based metrics. PEC evaluates the story plan, so it is shared by both backbones and not applicable to Direct I2V (--). Bold indicates the best result in each column for each backbone.}
\label{tab:main_results}
\centering
\small
\setlength{\tabcolsep}{4.5pt}
\begin{adjustbox}{max width=\textwidth}
\begin{tabular}{lccccccccc}
\toprule
& \multicolumn{2}{c}{Narrative $\uparrow$} & \multicolumn{3}{c}{Coherence $\uparrow$} & \multicolumn{3}{c}{Consistency $\uparrow$} & \multicolumn{1}{c}{Overall $\uparrow$} \\
\cmidrule(lr){2-3}\cmidrule(lr){4-6}\cmidrule(lr){7-9}\cmidrule(lr){10-10}
Method & PEC & ECS & APR & SPS & LAR & GPC & MIR & LCS & Avg \\
\midrule
\multicolumn{10}{l}{\textit{Veo~3.1 backbone}} \\
Direct I2V & -- & 0.6225 & 0.5389 & 0.4406 & 0.8121 & 0.7342 & 0.3012 & 0.2347 & 0.5263 \\
ViMax & 0.9052 & 0.9067 & 0.5500 & 0.5787 & 0.8684 & 0.8540 & 0.3850 & 0.2488 & 0.6274 \\
MovieAgent & 0.9694 & 0.8783 & 0.4333 & 0.5579 & 0.7889 & 0.6535 & 0.2508 & 0.2125 & 0.5393 \\
\textbf{\method (Ours)} & \textbf{0.9798} & \textbf{0.9225} & \textbf{0.9389} & \textbf{0.6681} & \textbf{0.9800} & \textbf{0.8971} & \textbf{0.4072} & \textbf{0.5691} & \textbf{0.7690} \\
\midrule
\multicolumn{10}{l}{\textit{Wan2.2-TI2V-5B backbone}} \\
Direct I2V & -- & 0.5850 & 0.6722 & 0.4653 & 0.7358 & 0.6866 & 0.2973 & 0.2272 & 0.5242 \\
ViMax & 0.9052 & 0.7750 & 0.5278 & 0.6024 & 0.8250 & 0.8617 & 0.3926 & 0.2633 & 0.6068 \\
MovieAgent & 0.9694 & 0.8350 & 0.3667 & 0.6129 & 0.7688 & 0.6925 & 0.2849 & 0.2366 & 0.5425 \\
\textbf{\method (Ours)} & \textbf{0.9798} & \textbf{0.8733} & \textbf{0.8444} & \textbf{0.6219} & \textbf{0.9749} & \textbf{0.8820} & \textbf{0.4264} & \textbf{0.5372} & \textbf{0.7372} \\
\bottomrule
\end{tabular}
\end{adjustbox}
\end{table}

\paragraph{Quantitative results.}
As shown in Table~\ref{tab:main_results}, \method{} achieves the best score on all seven video-based metrics with both backbones, together with the highest PEC among the methods that produce a plan. Its Avg reaches 0.7690 with Veo~3.1 and 0.7372 with Wan2.2-TI2V-5B, exceeding that of ViMax, the strongest baseline, by 0.1416 and 0.1304, respectively, and the Avg ranking of all methods is identical across the two backbones. Replacing Veo~3.1 with the open-source Wan2.2-TI2V-5B lowers ECS for every method, most sharply for ViMax (from 0.9067 to 0.7750), whereas the ECS margin of \method{} over the strongest baseline grows from 0.0158 to 0.0383, suggesting that its explicit render criteria and repair loop matter more when the backbone is weaker.

\paragraph{Comparison with baselines.}
Direct I2V renders every shot independently from a prompt that contains neither events nor states, and it has the lowest Avg, ECS, and SPS with both backbones. The agentic baselines plan the story and raise ECS and SPS substantially, e.g., ECS from 0.6225 to 0.9067 for ViMax with Veo~3.1, but they barely improve anchor persistence across cuts: their APR stays at or below 0.5500 and, with Wan2.2-TI2V-5B, falls below that of Direct I2V (0.6722), whose prompts name the opening anchor of each shot. Moreover, the LCS of every baseline lies between 0.21 and 0.27, and MovieAgent has the lowest GPC and MIR among the agentic methods. In contrast, \method{} gains most on cut- and location-level structure, raising APR to 0.9389 and 0.8444, LAR to 0.9800 and 0.9749, and LCS to 0.5691 and 0.5372, while also leading in ECS, SPS, GPC, and MIR by margins between 0.0090 and 0.0894. We attribute these gains to the continuity gate, which carries compatible anchors across cuts, to the canonical character and scene references, which fix recurring identities and environments, and to the evaluation-guided repair loop, which corrects residual violations; the ablations in Table~\ref{tab:ablation_results} support this attribution. SPS nevertheless remains below 0.67 for all methods, leaving irreversible state progression as an open challenge.

\paragraph{Qualitative comparison.}
Figure~\ref{fig:teaser} compares the three agentic methods on the same story, which requires the states of the backpack and notebook to persist across revisits to the classroom and the locker room. ViMax places the backpack in the locker before it is stored and omits the storage action, and both ViMax and MovieAgent drift in scene layout or character appearance across shots. In contrast, \method{} keeps the notebook on the desk until it is retrieved, leaves the backpack in the locker during the classroom revisit, and reunites the two objects in the final shot, while preserving the character and the recurring environments.

\subsection{Ablation Study}
\begin{figure}[!t]
    \centering
    \includegraphics[width=0.75\textwidth]{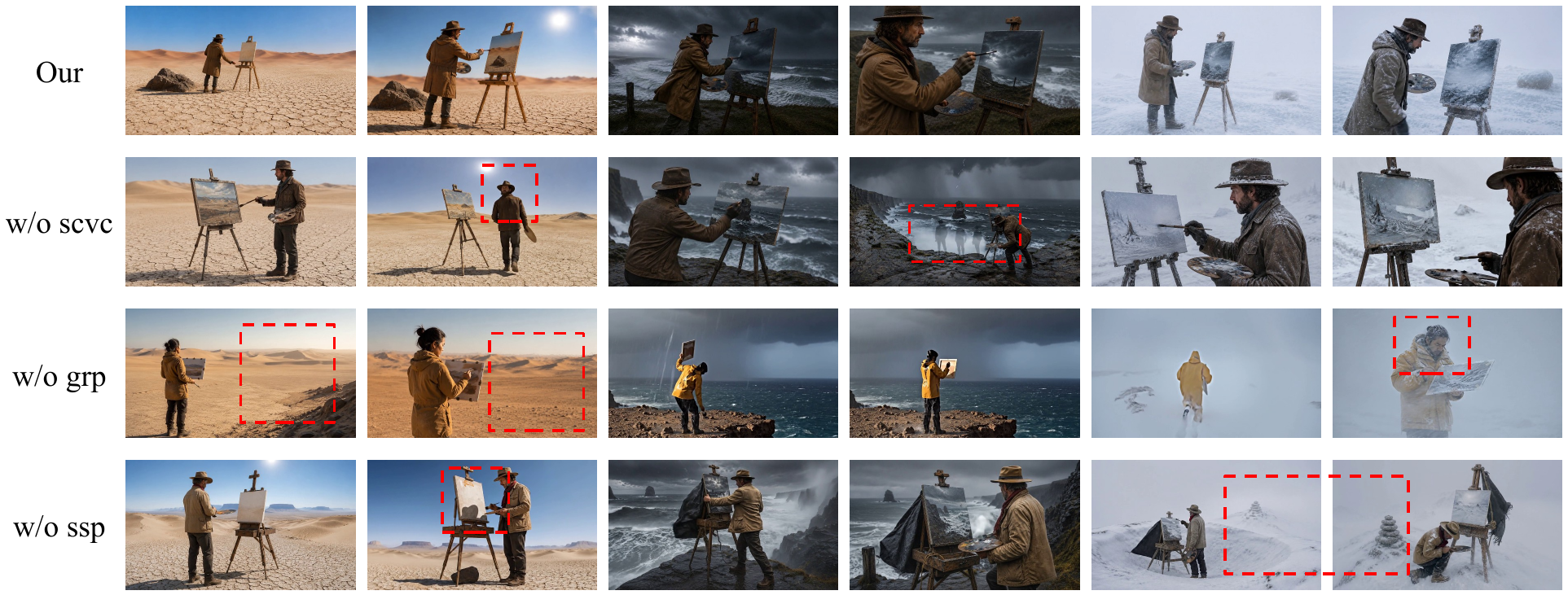}
    \caption{Qualitative ablation on a single story. From top to bottom, the rows show the full \method{}, \texttt{w/o~SCVC}, \texttt{w/o~GRP}, and \texttt{w/o~SSP}. Dashed boxes mark representative errors.}
    \label{fig:ablation}
\end{figure}

\begin{table}[!t]
\caption{Component ablations and planner sensitivity with the Veo~3.1 backbone. The Gemini-3.5-Flash planner row replaces only the GPT-5.5 planner of the full system. PEC is not applicable to \texttt{w/o~SSP} (--), which removes the story-planning stage of \method{}, and thus produces no per-shot story plan. Avg is defined as in Table~\ref{tab:main_results}. Bold marks the best score in each column.}
\label{tab:ablation_results}
\centering
\small
\setlength{\tabcolsep}{4.5pt}
\begin{adjustbox}{max width=\textwidth}
\begin{tabular}{lccccccccc}
\toprule
& \multicolumn{2}{c}{Narrative $\uparrow$} & \multicolumn{3}{c}{Coherence $\uparrow$} & \multicolumn{3}{c}{Consistency $\uparrow$} & \multicolumn{1}{c}{Overall $\uparrow$} \\
\cmidrule(lr){2-3}\cmidrule(lr){4-6}\cmidrule(lr){7-9}\cmidrule(lr){10-10}
Configuration & PEC & ECS & APR & SPS & LAR & GPC & MIR & LCS & Avg \\
\midrule
\texttt{w/o~SSP} & -- & 0.9108 & 0.9111 & 0.6374 & 0.9750 & 0.8800 & 0.4055 & 0.5635 & 0.7548 \\
\texttt{w/o~GRP} & 0.9684 & 0.9176 & 0.9153 & 0.5956 & 0.9529 & 0.7897 & 0.3845 & 0.4280 & 0.7119 \\
\texttt{w/o~SCVC} & 0.9766 & 0.8796 & 0.8264 & 0.6648 & 0.9750 & 0.8486 & 0.3653 & 0.5157 & 0.7251 \\
\midrule
Gemini-3.5-Flash planner & 0.8872 & 0.8898 & 0.8889 & 0.6643 & 0.9700 & 0.8087 & \textbf{0.4179} & 0.5103 & 0.7357 \\
\midrule
\textbf{\method{} (full)} & \textbf{0.9798} & \textbf{0.9225} & \textbf{0.9389} & \textbf{0.6681} & \textbf{0.9800} & \textbf{0.8971} & 0.4072 & \textbf{0.5691} & \textbf{0.7690} \\
\bottomrule
\end{tabular}
\end{adjustbox}
\end{table}

\paragraph{Variants.}
With the Veo~3.1 backbone, we remove one stage of \method{} at a time. \texttt{w/o~SSP} removes story-state propagation (Section~\ref{sec:story_state_propagation}) and infers the opening state of each shot independently. \texttt{w/o~GRP} removes grounded render planning (Section~\ref{sec:visual_grounding}); it retains state propagation but uses neither canonical references nor spatial grounding. \texttt{w/o~SCVC} removes continuity-aware generation (Section~\ref{sec:continuity_aware_generation}); each shot starts from a freshly generated opening frame, without tail reuse or evaluation-guided repair. To assess planner sensitivity, a separate variant replaces the GPT-5.5 planner with Gemini-3.5-Flash while keeping all other components and backends fixed (Appendix~\ref{app:baselines}).

\paragraph{Quantitative results.}
Table~\ref{tab:ablation_results} shows that removing any stage lowers Avg. \texttt{w/o~GRP} causes the largest drop, from 0.7690 to 0.7119, with GPC and LCS decreasing by 0.1074 and 0.1411, respectively. \texttt{w/o~SCVC} lowers Avg to 0.7251 and has the lowest ECS, APR, and MIR among the three component ablations. \texttt{w/o~SSP} causes the smallest drop in Avg (0.0142), with its largest decreases in SPS (0.0307) and APR (0.0278). SPS drops even further under \texttt{w/o~GRP} (by 0.0725), suggesting that visible state progression requires both propagated states and a grounded depiction of the state-bearing objects. These patterns are consistent with the intended roles of grounding, continuity control, and state propagation, respectively.

\paragraph{Planner sensitivity.}
With the Gemini-3.5-Flash planner, PEC and Avg decrease to 0.8872 and 0.7357, respectively, whereas MIR increases slightly from 0.4072 to 0.4179. This Avg still exceeds those of all Veo~3.1 baselines in Table~\ref{tab:main_results}, including ViMax and MovieAgent, even though both plan with GPT-5.5. The Avg advantage of \method{} therefore holds with both tested planners.

\paragraph{Qualitative comparison.}
In the examples highlighted in Figure~\ref{fig:ablation}, \texttt{w/o~SCVC} exhibits anatomical artifacts and ghosting, \texttt{w/o~GRP} changes the character's appearance and the scene content, and \texttt{w/o~SSP} introduces discontinuities in the canvas content and object placement. The full system maintains a more consistent appearance and coherent progression of the canvas content.

\section{Conclusion}
% We introduced \method, a state-grounded agentic framework for long-form video storytelling. By propagating explicit story states, grounding render plans, and repairing local inconsistencies, \method preserves causal progression and prevents visual errors from becoming narrative facts. On a new 60-story benchmark, \method consistently outperforms representative baselines in narrative realization, cross-shot coherence, and visual consistency.
We introduced \method, a state-grounded agentic framework that separates authoritative semantic plans from fallible visual observations in long-form video storytelling. By propagating explicit story states, grounding render plans in canonical references, and repairing local inconsistencies, \method preserves causal progression and prevents visual errors from becoming narrative facts. On a new 60-story benchmark, it consistently outperforms representative baselines in narrative realization, cross-shot coherence, and visual consistency, with the largest gains in anchor persistence and lighting coherence. Remaining challenges include irreversible state progression and longer stories with more locations and characters.

%%%%%%%%%%%%%%%%%%%%%%%%%%%%%%%% Bibliography %%%%%%%%%%%%%%%%%%%%%%%%%%%%%%%%%%
\bibliography{iclr2027_conference}
\bibliographystyle{iclr2027_conference}

\appendix
\providecommand{\nolinkurl}[1]{\url{#1}}

\section{Appendix Overview}
\label{app:overview}

The appendix is organized to support reproducibility. Appendix~\ref{app:method} specifies the complete state-grounded pipeline, including its structured contracts, prompt interfaces, and implementation settings, together with the protocols for the baselines and ablations. Appendix~\ref{app:benchmark} describes the benchmark design, the suite-specific annotations, per-suite statistics, the complete story index, and example briefs. Appendix~\ref{app:evaluation} gives the complete protocol for all eight metrics, including frame sampling, judge interfaces, aggregation rules, and thresholds. Appendix~\ref{app:limitations} discusses the limitations of the method and of the evaluation.

\section{Complete Method Specification}
\label{app:method}

This section specifies the semantic contracts, generation policy, and implementation settings of \method{}, together with the protocols for the baselines and ablations. The notation follows Section~\ref{sec:method}: generated pixels are observations, whereas the propagated story state is the sole authority on narrative facts.

\subsection{End-to-End Procedure}
\label{app:end_to_end}

The framework first constructs and validates the authoritative semantic plan $\mathcal{P}_{\mathrm{s}}$. Grounded references and shot contexts are then compiled into the executable render plan $\mathcal{P}_{\mathrm{r}}$. These two stages are completed before video synthesis, so the intended state trajectory remains fixed throughout generation.

Shots are rendered in temporal order because the tail of shot $i-1$ can provide useful visual evidence for shot $i$. The continuity gate selects reuse, reference, or fresh mode without modifying the semantic plan. Each candidate is evaluated against the fixed phase-tagged criteria, and failed or unknown criteria guide bounded local repair with $T=3$. The accepted shots are concatenated in their planned order.

\subsection{Authoritative Semantic Plan}
\label{app:semantic_plan}

\paragraph{Identifiers and catalogs.}
The planner first assigns stable string identifiers to recurring characters, props, and environments. An entity identifier denotes the same story participant everywhere; a state-specific visual asset is selected through the entity's attributes rather than by creating a new entity. The environment catalog similarly separates the identity of a place from the camera view used in a particular shot.

The semantic plan retains the story request and global requirements; stable entity and environment catalogs; placements and story-relevant attributes; ordered events and their effects; spatial intents; and phase-specific requirements. Free-form descriptions support generation, while cross-shot operations use stable IDs, typed placements, ordered effects, and explicit criteria. Each shot binds its events and presentation intent to one environment and one validated state transition.

\paragraph{Typed placements.}
The placement vocabulary is
\begin{equation}
\begin{aligned}
\mathcal{D}_{\mathrm{place}}=\{&\texttt{in\_scene\_zone},\texttt{on\_surface},\\
&\texttt{in\_container},\texttt{held\_by},\\
&\texttt{attached\_to},\texttt{offscreen}\}.
\end{aligned}
\label{eq:placement_domain}
\end{equation}
Each relation is type checked. For example, \texttt{held\_by} must target a character, \texttt{in\_container} must target a container-like prop, and a scene-zone placement must name a zone belonging to the active environment. The validator also rejects self-relations and cycles in the containment graph. These checks prevent syntactically valid plans from implying impossible ownership or containment chains.

\paragraph{State reduction and validation.}
Let an event effect be a partial update $\delta_{i,j}$ to one entity record. For the ordered event list $\mathcal{A}_i=(a_{i,1},\ldots,a_{i,m_i})$, reduction is
\begin{equation}
\mathcal{S}_{i+1}=R(\mathcal{S}_i,\mathcal{A}_i)
=\delta_{i,m_i}\circ\cdots\circ\delta_{i,1}(\mathcal{S}_i).
\label{eq:appendix_reducer}
\end{equation}
An event with no effect is observational and leaves the state unchanged. After every nonempty update, the system checks that the referenced entity exists, attributes lie in their declared domains, the placement target has a compatible type, and the placement graph is acyclic. The full trajectory is recomputed from $\mathcal{S}_1$; planner-supplied later states are treated as assertions to validate, not independent facts.

\paragraph{Illustrative transition.}
Suppose a notebook begins \texttt{on\_surface(desk)} and a backpack begins \texttt{held\_by(student)}. An event ``the student stores the backpack in the locker'' has the effect
\begin{equation}
\delta(\texttt{backpack})=\texttt{in\_container(locker)}.
\end{equation}
The next classroom shot therefore opens with the notebook still on the desk and the backpack in the locker, even if the preceding generated clip accidentally shows a backpack-like object in frame. A later retrieval event can update the notebook to \texttt{held\_by(student)} without revisiting or reinterpreting those pixels.

\subsection{Grounding and Render-Contract Compilation}
\label{app:grounding}

\paragraph{Canonical asset library.}
For a recurring character, the library stores a neutral full-body identity sheet with front, side, and rear evidence. For a visually stateful prop, it stores one reference per planned appearance state, and for a recurring environment it stores a $2{:}1$ panorama. Identity and state labels are inherited from the semantic plan. Assets are screened for correspondence to the canonical description, absence of unrequested text or people, and technical usability. A rejected asset is regenerated without changing its semantic specification.

The panorama is queried through six canonical probes---front, right, back, left, up, and down---to expose layout failures before shot generation. Each shot then receives an action-relevant perspective crop or synthesized view, with visible landmarks and an explicit camera specification. The same panorama anchors all views of an environment; the view is evidence for rendering and is not a new environment record.

\paragraph{Asset binding.}
For every entity visible in shot $i$, the compiler resolves a binding
\begin{equation}
b_i(e)=\bigl(e,\mathcal{S}_i(e),\ell(e,\mathcal{S}_i(e))\bigr),
\label{eq:asset_binding}
\end{equation}
where $\ell$ chooses the canonical identity/state asset compatible with the opening state. End-state assets may additionally be supplied when a visible transition changes appearance. This explicit lookup avoids asking a text prompt to rediscover which visual variant belongs to the current narrative state.

\paragraph{Phase-tagged criteria.}
The compiler normalizes each criterion to the tuple
\begin{equation}
k=(\text{id},\text{phase},\text{priority},\text{statement},
\text{evidence type}),
\end{equation}
where phase is \texttt{start}, \texttt{motion}, \texttt{end}, or \texttt{always}; priority distinguishes requirements from preferences. Start criteria are compiled from visible facts in $\mathcal{S}_i$, motion criteria from $\mathcal{A}_i$, and end criteria from visible facts in $\mathcal{S}_{i+1}$. Always-on criteria cover identity, environment, and technical constraints. A criterion is written as a positive, visually testable statement. For instance, ``the locker is the backpack's location'' becomes ``at the opening, no backpack is visible in the classroom; the locker remains the referenced storage location,'' rather than a request to infer an unseen symbolic relation.

The complete grounded context is
\begin{equation}
\begin{aligned}
\mathcal{U}_i=(&\text{scene view},\text{camera},\text{bindings},\\
&\text{landmarks},\text{exclusions}),
\end{aligned}
\end{equation}
and $r_i=(\mathcal{S}_i,\mathcal{A}_i,\mathcal{S}_{i+1},\mathcal{U}_i,\mathcal{K}_i)$ is serialized as structured data. The generator receives concise natural-language render instructions derived from this record, whereas the evaluator receives the canonical criteria themselves. Thus, paraphrasing in a generator prompt cannot silently alter the acceptance test.

\subsection{Continuity Gate and Local Repair}
\label{app:gate_repair}

\paragraph{Tail selection.}
The final decodable frame of the accepted previous shot is sampled after removing any terminal padding. If decoding fails or the frame is technically unusable, the tail is marked unavailable and the gate returns fresh mode. Otherwise, a VLM evaluates the frame only against the next shot's start/always criteria and returns PASS, FAIL, or UNKNOWN with criterion-linked evidence.

\paragraph{Deterministic mode resolver.}
Let $q_i$ be the number of failed or unknown hard opening criteria. The resolver follows the policy below.
\begin{enumerate}
    \item If no usable tail exists, choose \emph{fresh}.
    \item If $q_i\leq1$, propose \emph{reuse}. The proposed opening is evaluated once more after all conditioning is assembled; any hard failure triggers fallback.
    \item If $q_i\geq2$ but compatible identity, wardrobe, environment, or viewpoint evidence can be isolated, choose \emph{reference}. The composition prompt lists both the evidence to preserve and the incompatible content to exclude.
    \item Otherwise choose \emph{fresh} and condition only on the canonical library and current render contract.
\end{enumerate}
The threshold is deliberately conservative: a single uncertain observation may still preserve cut continuity, whereas multiple conflicts indicate that directly copying the frame risks propagating an error. The gate never writes to $\mathcal{S}_i$ or $\mathcal{S}_{i+1}$.

\paragraph{Opening validation.}
In reuse mode, the tail is the candidate opening image. In reference and fresh modes, an image model composes a new candidate using the shot view, state-appropriate entity references, start criteria, and exclusions. The candidate must satisfy every hard start criterion and pass technical checks before video synthesis. Fallback is monotone: \emph{reuse}$\rightarrow$\emph{reference}$\rightarrow$\emph{fresh}; a later stage cannot reintroduce rejected tail content.

\paragraph{Criterion-wise evaluation.}
For candidate $v_i^{(t)}$, the evaluator is shown endpoint and uniformly sampled interior frames, the fixed criterion list, and no candidate from another method. Its structured response contains one label and a short visual rationale per criterion. UNKNOWN is not converted to PASS; it remains a repair target and counts as a hard violation when the criterion is required.

Only failed or unknown canonical statements are appended to the next generation request. Successful requirements, the camera specification, asset bindings, and the semantic transition remain unchanged. This makes the loop local and plan preserving: it resamples an execution of $r_i$ rather than replanning the story.

\paragraph{Candidate selection.}
If no candidate satisfies all required criteria within the budget, candidates are ordered lexicographically by
\begin{equation}
\mathrm{Rank}(v;r_i)=
\bigl(n_{\mathrm{req}},n_{\mathrm{unk}},L_{\mathrm{prio}},
L_{\mathrm{cont}},L_{\mathrm{pref}},L_{\mathrm{tech}}\bigr),
\label{eq:candidate_rank}
\end{equation}
where $n_{\mathrm{req}}$ is the number of failed hard requirements, $n_{\mathrm{unk}}$ is the number of unknown judgments, $L_{\mathrm{prio}}$ weights remaining failures by criterion priority, and the last three terms measure continuity, preference, and technical losses. This order prevents a visually polished candidate from outranking one that realizes more of the required state transition. Ties are resolved by earlier generation order, yielding deterministic selection from a fixed candidate set.

\paragraph{Audit record.}
For every shot, we retain the immutable input contract, chosen mode, admissible reference hashes, opening-frame judgments, provider parameters, each candidate, all criterion decisions, repair text, final rank, and selection reason.

\subsection{Prompt Interfaces}
\label{app:prompts}

Listings~\ref{lst:planner_prompt}--\ref{lst:evaluator_prompt} show the operative prompt contracts in condensed form. Descriptive fields and story-specific values are inserted as structured JSON.

\begin{listing}[t]
\begin{lstlisting}[numbers=none]
SYSTEM: Convert the request into an authoritative story plan.
Return only the supplied JSON schema.
- Use stable IDs for every recurring entity and environment.
- Give the complete initial placement and relevant attributes.
- For each shot, list events in narrative order and attach only
  effects that change the world state.
- State spatial intent and positive, visually testable criteria.
- Never infer a later state from hypothetical generated pixels.
USER: {request, target_shots, style, schema}
\end{lstlisting}
\caption{Condensed semantic-planning prompt contract.}
\label{lst:planner_prompt}
\end{listing}

\begin{listing}[t]
\begin{lstlisting}[numbers=none]
TASK: Produce the opening image for this fixed render contract.
MUST SHOW: {hard_start_criteria}
PRESERVE: {compatible evidence}
DO NOT COPY: {tail conflicts and exclusions}
CAMERA AND LAYOUT: {grounded_view_and_camera}
REFERENCES: {state_appropriate_asset_bindings}
Do not add unlisted characters, props, text, logos, or watermarks.

VIDEO TASK: Animate the accepted opening through the ordered events.
The beginning, motion, and ending must satisfy their tagged criteria.
Keep identity, wardrobe, object state, layout, and screen direction.
REPAIR TARGETS: {failed or unknown canonical criteria}
\end{lstlisting}
\caption{Condensed opening-image and shot-video prompt interfaces. Empty preserve/exclusion fields are omitted.}
\label{lst:generation_prompt}
\end{listing}

\begin{listing}[t]
\begin{lstlisting}[numbers=none]
SYSTEM: Judge only visible evidence in the supplied media.
For every criterion return exactly one of PASS, FAIL, UNKNOWN.
PASS requires clear evidence; absence or ambiguity is not PASS.
Do not repair, rewrite, or add requirements.
Return JSON:
{criterion_results: [{id, label, evidence}],
 technical_validity, summary}
INPUT: {fixed_render_contract, criterion_list, sampled_frames}
\end{lstlisting}
\caption{Condensed criterion evaluator contract.}
\label{lst:evaluator_prompt}
\end{listing}

\subsection{Implementation and Generation Details}
\label{app:implementation}

\paragraph{Models and media settings.}
Table~\ref{tab:implementation} lists the generation and evaluation backends. Within each comparison, all methods render every shot with the same video backbone. Veo~3.1 produces $1280\!\times\!720$ shots at 24~fps with 144 frames (6~s). Wan2.2-TI2V-5B produces $1280\!\times\!704$ shots, the 720p resolution supported by the model, at 24~fps with 145 frames (approximately 6.04~s), because the model requires $4n{+}1$ frames. All API calls use structured outputs where available.

\begin{table}[t]
\caption{Generation and evaluation backends.}
\label{tab:implementation}
\centering
\small
\begin{tabular}{@{}l>{\raggedright\arraybackslash}p{0.62\linewidth}@{}}
\toprule
Component & Model/backend \\
\midrule
Planning & GPT-5.5~\citep{gpt55} for all LLM-based methods; Gemini-3.5-Flash~\citep{gemini35flash} in the planner-sensitivity variant \\
Image generation & GPT-Image-2~\citep{gptimage2} \\
Video generation & Veo~3.1~\citep{veo} and Wan2.2-TI2V-5B~\citep{Wan} \\
In-loop evaluator of \method & Gemini-3.5-Flash \\
Metric judge & Gemini-3.5-Flash, temperature 0 \\
Frame features (ECS) & DINOv2-base~\citep{dinov2} \\
Geometry (GPC) & SIFT~\citep{sift} and USAC-MAGSAC~\citep{magsacpp} in OpenCV~\citep{opencv} \\
Person masking (GPC) & DeepLabV3--ResNet50~\citep{deeplabv3} \\
Face identity (MIR) & InsightFace \texttt{buffalo\_l}~\citep{scrfd,arcface} \\
\bottomrule
\end{tabular}
\end{table}

\paragraph{Technical video checks.}
Before semantic evaluation, each candidate must be decodable, contain a non-empty frame sequence, have the expected width and height, and have a duration compatible with the request. Outputs that are entirely black or white, that are frozen because of a decoding failure, or whose output path is missing are rejected. A valid shot with little motion is not rejected solely because of low optical flow. No reported metric uses audio.

\subsection{Baseline and Ablation Protocols}
\label{app:baselines}

All methods receive the same benchmark brief, namely the story idea with its shot-to-location plan, the visual style, the required entities, the forbidden content, and the delivery requirements; no method has access to the annotation sheets. All LLM-based methods plan with GPT-5.5, all methods generate images with GPT-Image-2, and within each comparison all methods use the same video backbone. Except for PEC, which reads the per-shot plan text, all metrics are computed only from the exported shot videos and the frozen annotations; neither render prompts nor method names are shown to the metric judge.

\paragraph{Direct I2V.}
This planner-free baseline makes no language-model call. For each shot, a deterministic parser extracts the planned location and, when the brief specifies it, the entity on which the shot opens, and composes a prompt that also contains the required entities and the visual style. GPT-Image-2 generates the opening keyframe of each shot independently at the resolution of the backbone, the backbone renders each shot from its keyframe and prompt, and the shots are concatenated in plan order. The baseline uses no previous tail, canonical reference, state table, or evaluator-guided repair. Because it has no story-planning module and produces no plan, PEC is not applicable to it.

\paragraph{ViMax.}
We retain the public ViMax planning and storyboarding pipeline and its native cross-scene orchestration, and use GPT-5.5, GPT-Image-2, and the evaluated video backbone as its language, image, and video backends. Its scene plans are expanded into the ten requested shots, and its storyboard frames initialize image-to-video generation.

\paragraph{MovieAgent.}
We retain MovieAgent's multi-agent decomposition, character and scene asset construction, storyboard ordering, and region-of-interest (ROI) control interface, and replace its original language, image, and video backends with GPT-5.5, GPT-Image-2, and the evaluated video backbone. Because MovieAgent requires a character bank that the benchmark does not provide, we synthesize the bank for each story from its brief with the same GPT-5.5 and GPT-Image-2 backends. ROI boxes and layout controls are serialized into the corresponding image and video prompts.

\paragraph{Ablations and planner sensitivity.}
All ablations are derived from the same implementation of \method{}, and each removes a single stage. \texttt{w/o~SSP} infers the complete opening state of every shot independently from the story idea, the static catalogs, the initial world state, and the current shot, so adjacent shots never share a propagated state. Without this story-planning stage, the variant produces no per-shot plan of events, so PEC, which reads this plan, is not applicable to it. \texttt{w/o~GRP} retains propagated states and state-conditioned continuity but never generates, binds, or evaluates visual references or spatial evidence, so render planning is text-only. \texttt{w/o~SCVC} starts every shot from a newly generated opening frame, without tail extraction or reuse; its evaluators still run, but their feedback never triggers another generation attempt. The planner-sensitivity variant replaces GPT-5.5 with Gemini-3.5-Flash for planning and keeps all other backends fixed.

\section{Benchmark Details}
\label{app:benchmark}

\subsection{Design and Construction}

The benchmark contains 60 stories organized into three diagnostic suites of 20 stories each: N20 for narrative realization, T20 for cross-shot coherence, and C20 for visual consistency. Each suite covers five setting categories with four stories per category (Table~\ref{tab:benchmark_index}), yielding 50 distinct settings; ten settings appear in both N20 and T20, with different briefs and annotations. Table~\ref{tab:benchmark_stats} summarizes the three suites. Every story consists of a brief and an annotation sheet.

\paragraph{Brief.}
The brief is the only input given to a method. It is written in English and contains the story idea, a target of 10 shots, the visual style, five required recurring entities, a list of forbidden content, and the delivery requirements ($1280\!\times\!720$ resolution, 24~fps, shot durations of 4, 6, or 8~s, and no audio). The idea names the cast and the two locations and states the shot-to-location plan explicitly, e.g., ``shots 1--3 in the bakehouse, shots 4--5 in the flour store''. All briefs forbid on-screen text, logos and watermarks, split screens, frozen frames, more than one location within a shot, montage or time-lapse compression, speed ramping, flashback treatments, a visible film crew, direct address to the camera, and aerial viewpoints. T20 briefs additionally forbid extreme close-ups of isolated body parts, and C20 briefs forbid any second person, including bystanders, reflections, and depictions of people.

\paragraph{Annotation sheet.}
Each brief is paired with an annotation sheet that is never exposed to any method. The sheet is organized into typed blocks, including the location plan with location descriptions, the cast with the appearance strings specified in the brief, the key props, and the suite-specific blocks described below. Every block is frozen under a content hash, and the sheets were authored together with the briefs without consulting any system output.

\subsection{Suite Design}

\paragraph{N20: narrative realization.}
Each story has two named characters, four key props, and one of four location plans with five or six location changes. The brief lists ten events that must happen on screen, each visible as a change rather than as a person standing near an object. The annotation records the ten required events with their \emph{before} and \emph{after} visual predicates, together with six distractor events drawn from other N20 stories, which are used to measure false-positive judgments.

\paragraph{T20: cross-shot coherence.}
Each story has two named characters and follows one of four location plans with three or four location changes. For each of the nine cuts, the brief names one large, high-contrast anchor entity that must be visible at the end of the outgoing shot and at the start of the incoming shot, e.g., ``Shot 3 opens on the red-painted sack barrow in the store, and the shot before it ends on the same red barrow, held in frame long enough to read''. The annotation records the corresponding presence proposition for all 180 cuts, 71 of which involve a change of location. Each story also contains three monotone, i.e., irreversible, state changes (60 in total), such as a proving rack that is wheeled clear of the oven mouth and never returns. The brief lists these changes as jobs that may be placed in any shot but must never be undone, and the annotation records mutually exclusive \emph{before} and \emph{after} propositions together with the location at which each state is visible. For every change, the location of the state is planned for shots both before and after the reference position of the change.

\paragraph{C20: visual consistency.}
Each story features a single character and follows one of four location plans with five or six location changes, so that both locations are revisited. The brief requires a revisited room to keep the same walls, fittings, and floor, and asks for the character's face and clothing to remain legible in every shot. The plans contain 415 same-location shot pairs, 340 of which are non-adjacent. Lighting annotations are derived solely from the visual style of each brief: 13 stories specify a directional contrast between the two locations (11 in luminance and 2 in color temperature), the remaining 7 specify a single look, and no story requests a lighting change within a location.

\begin{table}[t]
\caption{Benchmark statistics per suite. MIR is computed on all 60 stories.}
\label{tab:benchmark_stats}
\centering
\small
\begin{tabular}{@{}lccc@{}}
\toprule
 & N20 & T20 & C20 \\
\midrule
Stories / shots & 20 / 200 & 20 / 200 & 20 / 200 \\
Locations per story & 2 & 2 & 2 \\
Location changes per story & 5--6 & 3--4 & 5--6 \\
Named characters per story & 2 & 2 & 1 \\
Required entities per brief & 5 & 5 & 5 \\
Forbidden-content items per brief & 11 & 12 & 15 \\
Required / distractor events & 200 / 120 & -- & -- \\
Anchored cuts & -- & 180 & -- \\
Monotone state changes & -- & 60 & -- \\
Same-location shot pairs & -- & -- & 415 \\
Declared lighting contrasts & -- & -- & 13 \\
Metrics & PEC, ECS & APR, SPS, LAR & GPC, LCS \\
\bottomrule
\end{tabular}
\end{table}

\begin{table}[t]
\caption{Complete story index by setting category and suite.}
\label{tab:benchmark_index}
\centering
\scriptsize
\setlength{\tabcolsep}{4pt}
% Standard p columns and explicit row endings keep this table independent of array.
\begin{tabular}{@{}p{0.1\linewidth}p{0.276\linewidth}p{0.276\linewidth}p{0.276\linewidth}@{}}
\toprule
\raggedright Category & \raggedright N20 & \raggedright T20 & \raggedright C20 \tabularnewline
\midrule
\raggedright Store & \raggedright Greenhouse Vent Crank; Icehouse Block Tongs; Kiln Shelf Load; Seedhouse Drill & \raggedright Bonded Store Seal; Greenhouse Vent Crank; Icehouse Block Tongs; Kiln Shelf Load & \raggedright Kiln Watcher; Orchard Grafter; Tide Recorder; Wheel Turner \tabularnewline
\midrule
\raggedright Food & \raggedright Bakehouse Peel Rack; Brewery Mash Rake; Cheeseroom Turning; Dairy Churn Belt & \raggedright Bakehouse Peel Rack; Brewery Mash Rake; Dairy Churn Belt; Saltworks Pan Rake & \raggedright Hop Picker; Press House; Salt Raker; Smoke Curer \tabularnewline
\midrule
\raggedright Transit & \raggedright Cablecar Gripman; Drydock Caisson; Lockgate Paddle; Tollhouse Barrier & \raggedright Ferry Slip Ramp; Funicular Haul Room; Lockgate Paddle; Tramshed Pit Road & \raggedright Bridge Keeper; Diver Tender; Net Braider; Slipway Greaser \tabularnewline
\midrule
\raggedright Workshop & \raggedright Cooperage Hoop; Forge Quench Tank; Ropewalk Traveller; Sailloft Bolt Rope & \raggedright Bindery Nipping Press; Forge Quench Tank; Glassworks Annealing Lehr; Sailloft Bolt Rope & \raggedright Furnace Charger; Glass Gatherer; Lock Fitter; Press Setter \tabularnewline
\midrule
\raggedright Control Room & \raggedright Observatory Dome; Powerhouse Switchboard; Pumproom Telegraph; Signal Lamp Room & \raggedright Lighthouse Watch Room; Pumphouse Governor; Signal Lamp Room; Telegraph Relay Room & \raggedright Bell Ringer; Gauge Reader; Lamp Trimmer; Signal Fitter \tabularnewline
\bottomrule
\end{tabular}
\end{table}

\subsection{Example Briefs}
\label{app:example_briefs}

The following abridged excerpts from the story ideas of one story per suite illustrate the brief format; omissions are marked by [\dots].

\paragraph{N20, Bakehouse Peel Rack.}
\begin{quote}\small
Ewa (baker, white cotton jacket, blue-checked head-cloth) and Tam (bakehouse hand, grey collarless shirt, canvas back-brace). Ten shots and two places: the bakehouse and the flour store. The plan is shots 1-3 in the bakehouse, shots 4-5 in the flour store, shot 6 in the bakehouse, shot 7 in the flour store, shots 8-9 in the bakehouse, shot 10 in the flour store. Every one of these ten things must actually happen on screen, and each one has to be visible as a CHANGE [\dots]. 1) Tam barrows a fresh sack in from the store on the red barrow. 2) Ewa weighs flour out on the hanging scales. [\dots] 10) Ewa leans the blue-handled peel back by the oven. Some of the order matters: [\dots]. Where the rest of the steps fall across the ten shots is yours to decide.
\end{quote}

\paragraph{T20, Bakehouse Peel Rack.}
\begin{quote}\small
Ewa bakes and Tam fetches from the store. [\dots] The jobs to be done, in whatever order the plan decides: wheel the steel proving rack clear of the oven mouth; raise the black oven door on its chain and leave it up; set the loaves with the long blue-handled peel; barrow a fresh sack through from the store; weigh the flour out on the hanging scales. Ten shots and only two places: the bakehouse and the flour store. The plan is shot 1 in the bakehouse, shots 2-3 in the flour store, shots 4-6 in the bakehouse, shots 7-8 in the flour store, shots 9-10 in the bakehouse. Every cut in this film is anchored on one thing that is plainly in frame on both sides of it, and the brief says which. [\dots] Shot 3 opens on the red-painted sack barrow in the store, and the shot before it ends on the same red barrow, held in frame long enough to read. [\dots] Where the jobs land across the ten shots is yours to decide, so long as each anchor is where the brief says it is at the cut, and none of the one-way changes below is ever undone.
\end{quote}

\paragraph{C20, Bell Ringer.}
\begin{quote}\small
Signe is tower keeper: a young woman with dark hair tied back, in a navy jumper and fingerless wool gloves. [\dots] Every one of the ten shots contains this one person and nobody else [\dots]. Ten shots and two places: the ringing chamber and the bell chamber above. The plan is shots 1-3 in the ringing chamber, shots 4-5 in the bell chamber above, shot 6 in the ringing chamber, shot 7 in the bell chamber above, shots 8-9 in the ringing chamber, shot 10 in the bell chamber above, and the film returns to each place more than once, so the same room has to look like the same room every time it comes back [\dots]. Frame the person clearly enough in every shot that their face and clothing can be read; where they stand in the room and the order of the work is yours.
\end{quote}

\section{Complete Evaluation Protocol}
\label{app:evaluation}

This section gives the complete protocol for the metrics summarized in Section~\ref{sec:evaluation_protocol}; symbols introduced here are local to this section. For a finite non-empty index set $\mathcal{I}$, let $\langle f_i\rangle_{i\in\mathcal{I}}\equiv |\mathcal{I}|^{-1}\sum_{i\in\mathcal{I}}f_i$ denote the mean over $\mathcal{I}$, and let $[N]=\{1,\ldots,N\}$, where $N=10$ is the number of planned shots per story, as in Section~\ref{sec:method}. We write $D$ for the duration of a shot in seconds and $\mathrm{linspace}(\alpha,\beta,m)$ for the $m$ evenly spaced values from $\alpha$ to $\beta$, inclusive.

\subsection{Common Protocol}

\paragraph{Inputs and judge.}
Except for PEC, which reads the per-shot plan text, every metric reads only the exported shot videos of a method, in planned order, together with the frozen annotation sheet of the story where one is required. The VLM judgments, namely those of PEC, ECS, APR, SPS, and LAR and the location gate of GPC, are made by Gemini-3.5-Flash~\citep{gemini35flash} at temperature 0, with structured JSON responses over a closed label vocabulary. The judge prompts instruct the model to report only the evidence it is shown, and APR, SPS, and LAR provide an explicit label for unreadable evidence so that the judge is not forced to guess. Except for the fixed 0.4~s windows at each cut in APR, frames are sampled at fixed fractions of each shot's duration, so the number of observations does not depend on clip length.

\paragraph{Aggregation and missing outputs.}
PEC, ECS, SPS, GPC, and LCS are first computed per story and then macro-averaged over delivered stories; APR and LAR pool their units (cuts and shots, respectively) over all delivered stories of a method; and MIR is averaged over all 60 stories, with an undelivered story scoring zero. Within a delivered story, a missing shot counts as showing no event in ECS, leaves its side of each adjacent cut undetermined in APR, is dropped from the probe sequences of SPS, has no readable location in LAR, and scores zero in every GPC pair that contains it. PEC, ECS, and GPC are computed in three independent evaluation runs with separate judge caches, and the reported value is the mean of the three run-level scores. Avg in Tables~\ref{tab:main_results} and~\ref{tab:ablation_results} is the unweighted mean of the seven video-based metrics; PEC is excluded from Avg because it evaluates plan text rather than video and is not applicable to methods without a story-planning module.

\subsection{Plan Event Coverage (PEC)}
\label{app:pec}

PEC evaluates the plan rather than the video and never reads a frame. It uses only the per-shot text that states what happens: the \texttt{action} fields of the beats of each \method{} shot, the \texttt{visual\_desc} field of each ViMax storyboard shot, and the \emph{Plot/Visual Description} field of each MovieAgent shot. The structured fields of \method{} plans (spatial intents, visible entities, preconditions, effects, operations, entity and scene catalogs, and plan invariants) and narrative-purpose statements are excluded, so that the richer structure of our plans does not enter the evaluation input; render prompts are not used for any method.

For each N20 story, the candidate list contains the ten required events and the six distractor events of the annotation sheet, shuffled and presented under opaque identifiers so that their source is not revealed. For each shot, the judge receives only the text of that shot and the candidate list, and labels each event as \emph{clear} if the text states that the event happens in this shot, \emph{partial} if the people or objects of the event appear in the text but the text does not state that the event happens here, and \emph{none} otherwise. The instructions note that a plan states an event in one plain clause, so a single clause stating the event suffices for \emph{clear}. Each \emph{clear} label must be supported by a quotation of at most 20 words copied from the shot text; the label is downgraded to \emph{partial} if the quotation is longer or if fewer than 60\% of its content words occur in the shot text.

Let $\mathcal{R}$ and $\mathcal{M}$ denote the sets of required and distractor events, respectively, and let $n_e$ be the number of shots in which event $e$ is labeled \emph{clear} after quotation verification. The credit of event $e$ is
\begin{equation}
r_e=
\begin{cases}
1, & 1\leq n_e\leq\lceil N/2\rceil,\\
\tfrac{1}{2}, & n_e>\lceil N/2\rceil,\\
\tfrac{1}{2}, & n_e=0\ \text{and some shot is labeled \textit{partial}},\\
0, & \text{otherwise}.
\end{cases}
\label{eq:event_credit}
\end{equation}
Thus, an event stated in more than five shots receives only partial credit. Writing $\bar r_{\mathcal{X}}\equiv\langle r_e\rangle_{e\in\mathcal{X}}$ for an event set $\mathcal{X}$, PEC is
\begin{equation}
\mathrm{PEC}=\frac{\bar r_{\mathcal{R}}-\bar r_{\mathcal{M}}}{1-\bar r_{\mathcal{M}}}.
\label{eq:pec}
\end{equation}
The distractor term acts as a per-story zero point that separates plan coverage from indiscriminate affirmative judgments. Because PEC evaluates story planning, it is not applicable to methods without a story-planning module: Direct I2V has no planner, and \texttt{w/o~SSP} removes story-state propagation and therefore produces no per-shot plan of events. Their PEC entries are marked with a dash (--) in Tables~\ref{tab:main_results} and~\ref{tab:ablation_results}. For the remaining methods, stories without plan text are excluded. PEC is computed per story, macro-averaged over stories, and averaged over three runs. \method{}, ViMax, and MovieAgent planned every story independently for each of the two backbones; because PEC does not depend on rendering, their reported values average the two planning runs. The PEC values of the remaining ablation and planner-sensitivity variants are computed from their Veo~3.1 runs.

\subsection{Event Completion Score (ECS)}
\label{app:ecs}

ECS uses the candidate list and label vocabulary of PEC but judges the video. Each shot is presented as a strip of six frames labeled F0--F5, sampled at $D\cdot\mathrm{linspace}(0.06,0.94,6)$, or at $D\cdot\mathrm{linspace}(0.12,0.88,6)$ if $D<6.7$, and the judge labels all candidates in a single call. The label \emph{clear} requires that the event visibly happens within the shot: the judge must cite two different frames $f_{\mathrm{before}}<f_{\mathrm{after}}$ and describe the change in at most 12 words. The label \emph{partial} applies when the people or objects of the event, or its finished state, are visible but the change cannot be demonstrated, and \emph{none} applies when none of them appears. A \emph{clear} label is downgraded to \emph{partial} if it violates this format or if the cosine distance between the DINOv2 embeddings~\citep{dinov2} of the two cited frames falls below the 5th percentile of the distances between adjacent sampled frames, so that a static depiction of the finished state does not count as completion.

The event credits $r_e$ follow Equation~\eqref{eq:event_credit}, applied to the video labels after evidence verification, so asserting every event in every shot is not rewarded. Let $y_{m,k}\in\{0,1\}$ indicate whether distractor $m$ receives the final label \emph{clear} in shot $k$, and let $\bar y_{\mathcal{M}}\equiv\langle y_{m,k}\rangle_{(m,k)\in\mathcal{M}\times[N]}$ denote the distractor false-positive rate. The per-story score is
\begin{equation}
\mathrm{ECS}=\bar r_{\mathcal{R}}-\bar y_{\mathcal{M}}.
\label{eq:ecs}
\end{equation}
ECS is macro-averaged over stories and averaged over three runs.

\subsection{Anchor Persistence Rate (APR)}
\label{app:apr}

For each planned cut of a T20 story, the evaluator composes a single image, called a card, with two labeled rows: BEFORE contains two stills spanning the final 0.4~s of the outgoing shot, and AFTER contains two stills spanning the first 0.4~s of the incoming shot. The judge receives the anchor proposition of the cut (``$X$ is in frame'') and labels each row as \emph{holds} (the named entity is visible in the described state), \emph{not\_holds} (it is visible in a different state), \emph{absent} (it does not appear), or \emph{unclear} (the stills are too blurred, dark, cropped, or occluded to decide). The instructions state that changes in camera angle, shot size, framing, lighting, or the positions of other people do not by themselves falsify the proposition, and that uncertain cases must be labeled \emph{unclear}. A side labeled \emph{holds} passes, a side labeled \emph{not\_holds} or \emph{absent} fails, and a side labeled \emph{unclear} is undetermined. A cut is satisfied if both sides pass, violated if either side fails, and undetermined otherwise.

Let $n_{\mathrm{sat}}$, $n_{\mathrm{vio}}$, and $n_{\mathrm{und}}$ be the numbers of satisfied, violated, and undetermined cuts, respectively, pooled over all delivered T20 stories. Then
\begin{equation}
\mathrm{APR}=\frac{n_{\mathrm{sat}}}{n_{\mathrm{sat}}+n_{\mathrm{vio}}+n_{\mathrm{und}}}.
\label{eq:apr}
\end{equation}
Undetermined cuts remain in the denominator and therefore count against the score.

\subsection{State Progression Score (SPS)}
\label{app:sps}

A T20 state is scored if it is monotone, has non-empty \emph{before} and \emph{after} propositions, and its location is planned for at least two shots; all 60 states satisfy these conditions. Every shot planned at the location of a state serves as a probe for that state, regardless of where the system places the change. For each probe, the judge sees one card whose top row, labeled MID, contains frames at 35\%, 50\%, and 65\% of the shot, and whose bottom row, labeled T-REF, contains frames at 5\% and 95\%. The judge labels the \emph{before} and \emph{after} propositions from the MID row only, using \emph{holds}, \emph{not\_holds}, \emph{absent}, or \emph{unclear}, and reports whether the named object changes state within the shot, i.e., whether both states appear in MID or a T-REF frame shows the state opposite to that in MID. The probe letter is $\mathrm{T}$ if such a transition is reported or both propositions hold, $\mathrm{B}$ if only the \emph{before} proposition holds, and $\mathrm{A}$ if only the \emph{after} proposition holds. Probes labeled \emph{absent} or \emph{unclear}, probes whose two propositions are both labeled \emph{not\_holds}, unparsable responses, and missing shots are dropped.

For a state $s$, the remaining probe letters, ordered by shot index, form the sequence $\sigma=(\sigma_1,\ldots,\sigma_n)$; states with an empty sequence are excluded. The state score is
\begin{equation}
\mathrm{SPS}_s
= \begin{cases}
    1, & \sigma\in\mathrm{B}^{+}\mathrm{T}^{*}\mathrm{A}^{+},\\
    0, & \sigma\in\mathrm{A}^{+}\cup\mathrm{B}^{+},\\
    \tfrac{1}{2}, & \sigma\in\mathrm{T}^{+},\\
    \rho(\sigma), & \text{otherwise},
\end{cases}
\label{eq:sps}
\end{equation}
where $^{+}$ and $^{*}$ denote one or more and zero or more repetitions, respectively, and $\rho(\sigma)$ is the best agreement with a single forward transition,
\begin{equation*}
\rho(\sigma)=\max_{1\leq k<n}\frac{1}{n}
\left(\sum_{i=1}^{k}\phi_{\mathrm{B}}(\sigma_i)
+\sum_{i=k+1}^{n}\phi_{\mathrm{A}}(\sigma_i)\right).
\end{equation*}
The functions $\phi_{\mathrm{B}}$ and $\phi_{\mathrm{A}}$ assign 1 to $\mathrm{B}$ and $\mathrm{A}$, respectively, 0.5 to $\mathrm{T}$, and 0 otherwise. Every sequence of length one falls under one of the first three cases, so $n\geq2$ whenever $\rho$ is used. SPS averages the state scores within each story and then macro-averages over stories.

\subsection{Location Adherence Rate (LAR)}
\label{app:lar}

LAR uses per-shot location records, which also serve the location gate of GPC. Each shot contributes an earlier and a later frame, at 20\% and 80\% of its duration, and up to six shots of the same story are shown as the columns of one card. The judge answers every column independently with one identifier from the location list of the story, which gives the name and description of each location, or with \emph{other} or \emph{cannot\_tell}. The instructions state that the same room seen from a different angle, with a different lens, or under different lighting remains the same location, and that \emph{other} applies only to a physically different place. Shots answered \emph{cannot\_tell}, undelivered shots, and shots whose decoding or judging failed have no readable location and are excluded.

Let $\mathcal{J}$ index all readable shots across the delivered T20 stories, and let $\hat{\ell}_j$ and $L_j$ denote the judged and planned locations of shot $j$. The pooled score is
\begin{equation}
\mathrm{LAR}=\left\langle\mathbf{1}[\hat{\ell}_j=L_j]\right\rangle_{j\in\mathcal{J}}.
\label{eq:lar}
\end{equation}
An answer of \emph{other} is readable and therefore counts as a mismatch.

\subsection{Geometric Place Consistency (GPC)}
\label{app:gpc}

\paragraph{Pairs and gate.}
For each C20 story, the candidate pairs are all shot pairs $(a,b)$ with $a<b$ that the frozen plan assigns to the same location, including adjacent pairs. The candidate set is therefore identical for all methods and cannot shrink when a method renders a location incorrectly. The gate $g_{ab}$ equals 1 if the location records of both shots (Appendix~\ref{app:lar}) are readable, differ from \emph{other}, and agree with each other, and 0 otherwise.

\paragraph{Geometric verification.}
Each shot contributes three frames, at 20\%, 50\%, and 80\% of its duration. People are removed with a DeepLabV3--ResNet50 person mask~\citep{deeplabv3} (probability threshold 0.5, dilated by 8 pixels), and at most 2000 SIFT keypoints~\citep{sift} are extracted from the remaining background. For each of the nine frame combinations of a pair, nearest-neighbor matches that pass Lowe's ratio test with ratio 0.75 are fitted with both a fundamental matrix and a homography using USAC-MAGSAC~\citep{magsacpp} with a 3-pixel threshold, confidence 0.999, and at most 10{,}000 iterations. Let $n_F$ be the number of fundamental-matrix inliers and $n_H$ the number of those that are also homography inliers. A combination is \emph{planar} if $n_H/n_F\geq0.85$, \emph{well spread} if the fundamental-matrix inliers occupy at least 3 cells of a $4\times4$ grid, and \emph{near-identical} if the median homography displacement is below 3\% of the image diagonal; it is \emph{degenerate} if it is planar but not both near-identical and well spread. Its admissible count is $n_F$ if it is well spread and not degenerate, and 0 otherwise, and $n_{ab}$ is the maximum admissible count over the nine combinations. A pair is degenerate, and is excluded from the average, if some combination has at least 8 fundamental-matrix inliers but $n_{ab}=0$. Together, these rules prevent a shared textured plane from being scored as a shared place.

\paragraph{Score.}
Let $\Pi$ be the set of non-degenerate candidate pairs of a story. With verification floor $F=8$ and saturation threshold $\tau=15$, define
\begin{equation}
\begin{aligned}
q_{ab} &= g_{ab}\mathbf{1}[n_{ab}\geq F]\min(1,n_{ab}/\tau),\\
\mathrm{GPC} &= \langle q_{ab}\rangle_{(a,b)\in\Pi}.
\end{aligned}
\label{eq:gpc}
\end{equation}
A pair that contains a missing shot has $q_{ab}=0$. GPC is macro-averaged over stories and averaged over three runs with independently judged location records.

\subsection{Minimum Identity Retention (MIR)}
\label{app:mir}

MIR is computed on all 60 stories. Four frames are sampled at evenly spaced indices of each shot, including its first and last frames. Faces are detected and embedded with the InsightFace \texttt{buffalo\_l} detector and recognizer~\citep{scrfd,arcface} at a detection size of $640\times640$, and detections with a confidence score below 0.55 are discarded. The $L_2$-normalized embeddings are processed in shot order, and each is greedily assigned to the existing cluster whose mean embedding has the highest cosine similarity with it, provided that this similarity is at least 0.45; otherwise, a new cluster is opened. Clusters with a single detection are discarded unless every cluster has a single detection. For each remaining cluster $c$, $\mathcal{Q}_c$ contains every pair of detections from different shots, and clusters without such pairs are ignored. The per-story score is
\begin{equation}
\mathrm{MIR}=\min_{c}\left\langle\cos(a,b)\right\rangle_{(a,b)\in\mathcal{Q}_c},
\label{eq:mir}
\end{equation}
where the minimum is taken over the clusters with non-empty $\mathcal{Q}_c$ and $\cos(a,b)$ is the cosine similarity between two face embeddings. Taking the minimum prevents a highly consistent cluster from compensating for a less consistent one. MIR is zero when fewer than two faces are detected or when no cluster spans two shots.

\subsection{Lighting Coherence Score (LCS)}
\label{app:lcs}

\paragraph{Lighting annotations.}
The lighting annotation of each C20 story partitions its shots into segments with lighting-condition identifiers. It is derived solely from the visual style of the brief: when the style explicitly contrasts the two locations, e.g., ``a cramped locker lit by one round port, against open grey daylight on deck'', the two locations receive different identifiers and a direction (darker or brighter, cooler or warmer); otherwise, all shots share one identifier. $\Pi_{\mathrm{same}}$ contains the non-adjacent shot pairs, i.e., pairs whose indices differ by at least 2, that share an identifier, and $\Pi_{\mathrm{diff}}$ contains the non-adjacent pairs with different identifiers; $\Pi_{\mathrm{diff}}$ is non-empty for 13 stories.

\paragraph{Descriptor and similarity.}
Each shot is sampled at eight frames, at $D\cdot\mathrm{linspace}(0.15,0.85,8)$. Every frame is converted to CIELAB with channels scaled to $[0,1]$, and its descriptor concatenates six scalars, namely the mean and standard deviation of lightness, the mean $a^{*}$ and $b^{*}$ offsets from 0.5, and the fractions of pixels with lightness above 0.8 and below 0.2, with a normalized 16-bin lightness histogram. Averaging over frames gives the 22-dimensional shot descriptor $u=(u^{\mathrm{s}},u^{\mathrm{h}})$, where $u^{\mathrm{s}}$ contains the six scalar features and $u^{\mathrm{h}}$ the histogram. The lighting similarity of shots $a$ and $b$ is
\begin{equation}
\mathrm{LS}(a,b)=1-\tfrac12\left[
\mathrm{clip}\left(\frac{\lVert u^{\mathrm{s}}_a-u^{\mathrm{s}}_b\rVert_1}{3},0,1\right)+
\mathrm{clip}\left(\frac{\lVert u^{\mathrm{h}}_a-u^{\mathrm{h}}_b\rVert_1}{2},0,1\right)
\right].
\label{eq:ls}
\end{equation}

\paragraph{Score.}
For every story with non-empty $\Pi_{\mathrm{same}}$ and $\Pi_{\mathrm{diff}}$, let $\bar L_{\mathrm{same}}$ and $\bar L_{\mathrm{diff}}$ denote the mean $\mathrm{LS}$ over the respective pair sets. With $z_\epsilon(x)=\operatorname{artanh}(\operatorname{clip}(x,\epsilon,1-\epsilon))$ and $[x]_+=\max(0,x)$, the per-story score is
\begin{equation}
\mathrm{LCS}=\tanh\!\left(\left[z_\epsilon(\bar L_{\mathrm{same}})-z_\epsilon(\bar L_{\mathrm{diff}})\right]_+\right).
\label{eq:lcs}
\end{equation}
Equivalently, $\mathrm{LCS}=[(x_{\mathrm{s}}-x_{\mathrm{d}})/(1-x_{\mathrm{s}}x_{\mathrm{d}})]_+$, where $x_{\mathrm{s}}$ and $x_{\mathrm{d}}$ are the clipped values of $\bar L_{\mathrm{same}}$ and $\bar L_{\mathrm{diff}}$; this Fisher $z$-transform therefore weights a given similarity gap more heavily when both similarities are close to 1. LCS is computed per story and macro-averaged over the 13 applicable stories. The clipping constant $\epsilon=0.02$, which prevents $\operatorname{artanh}$ from diverging at 1, was not tuned. Because the score contrasts two similarity levels, both a single frozen look for the whole film and lighting changes that ignore the plan yield values near zero.

\subsection{Scope of Metric Interpretation}

The eight metrics target distinct failure modes and should be read jointly rather than through Avg alone. PEC, ECS, APR, SPS, and LAR rely on a VLM judge, and GPC uses it for location gating; frozen propositions, closed label vocabularies, explicit labels for unreadable evidence, distractor events, and evidence verification limit this dependence, and PEC, ECS, and GPC are averaged over three runs. The judge is the same model as the in-loop evaluator of \method{} (Table~\ref{tab:implementation}), although the two receive different inputs: the evaluator checks the criteria compiled into \method{}'s own render contracts, whereas the judge checks frozen annotations that no method sees. Shared preferences of this model could nonetheless favor \method{} on the VLM-judged metrics; MIR and LCS involve no VLM and are unaffected by this overlap. PEC evaluates plan text rather than video, so it depends on how each system phrases its per-shot descriptions. APR verifies anchor presence rather than full physical continuity; SPS covers only the annotated irreversible states; LAR identifies the depicted location without checking its layout; GPC favors textured and approximately rigid backgrounds; MIR depends on face visibility and detector behavior; and LCS is defined only for the 13 stories that declare a lighting contrast, rewards a difference between the contrasted conditions without checking its annotated direction, and does not assess lighting changes within a location. Avg is an equal-weight summary rather than a calibrated preference score.

\section{Limitations}
\label{app:limitations}

\paragraph{Authoritative-plan errors.}
Separating semantics from pixels prevents visual errors from entering the story state, but it also makes planning mistakes persistent. If the initial entity catalog, the effect order, or the placement graph is wrong, later visual evidence cannot correct it autonomously. Schema checks catch structural contradictions but not every commonsense or cultural error. High-stakes use therefore requires human inspection or a separately validated replanning mechanism.

\paragraph{Finite visual evidence.}
Canonical references and a panorama constrain appearance but do not constitute a true 3D scene. Large viewpoint changes, severe occlusion, mirrors, transparent objects, deformable props, crowds, and fine hand--object contact remain difficult. A compatible endpoint also does not prove that the intervening motion is physically valid. The repair budget bounds cost and prevents runaway loops, but when it is exhausted, the selected candidate may still violate some criteria.

\paragraph{Evaluation coverage.}
The benchmark is in English, fictional, and visually oriented. Each of its 60 stories requests 10 shots over two locations in one of five categories of working environments, features one or two named characters, and has no audio. It therefore does not cover long narratives with many locations, crowded multi-person interaction, dialogue or audio continuity, lip synchronization, on-screen text, or interactive editing, and lighting contrasts are annotated only between locations. Future work should broaden the coverage of narratives, casts, and models and add human pairwise judgments.

\end{document}